\documentclass{article}
\usepackage{spconf,amsmath,graphicx}
\usepackage[utf8]{inputenc} 
\usepackage[T1]{fontenc}    
\usepackage{hyperref}       
\usepackage{url}            
\usepackage{booktabs}       
\usepackage{amsfonts}       
\usepackage{nicefrac}       
\usepackage{amsmath}
\usepackage{microtype}      

\usepackage{graphicx}       
\usepackage[noline,ruled]{algorithm2e} 
\usepackage{subcaption}

\usepackage{xcolor}

\title{U-Noise: Learnable Noise Masks for Interpretable Image Segmentation}
\name{
Teddy Koker $^{\star \mathparagraph}$\qquad
Fatemehsadat Mireshghallah$^{\dagger \mathparagraph}$\qquad
Tom Titcombe$^{\ddagger \mathparagraph}$\qquad
Georgios Kaissis$^{\mathsection \| \mathparagraph}$}
  
\address{
$^{\star}$Grid AI \qquad
$^{\dagger}$University of California San Diego \qquad
$^{\ddagger}$Tessella \\
$^{\mathsection}$Technical University of Munich \qquad
$^{\|}$Imperial College London \qquad
$^{\mathparagraph}$OpenMined
}

\begin{document}
%
\maketitle
\begin{abstract}
  Deep Neural Networks (DNNs) are widely used for decision making in a myriad of critical applications, ranging from medical to societal and even judicial. %
  Given the importance of these decisions, it is crucial for us to be able to interpret these models. 
  We introduce a new method for interpreting image segmentation models by learning regions of images in which noise can be applied without hindering downstream model performance.
  We apply this method to segmentation of the pancreas in CT scans, and qualitatively compare the quality of the method to existing explainability techniques, such as Grad-CAM and occlusion sensitivity. Additionally we show that, unlike other methods, our interpretability model can be quantitatively evaluated based on the downstream performance over obscured images.
\end{abstract}
\section{Introduction}

The ability to interpret the decisions made by predictive models helps to understand and evaluate model performance and has been a useful tool for identifying algorithmic biases~\cite{hendricks2018women}. In recent years, several methods for interpreting neural networks, which have typically been considered "black boxes", have been developed \cite{NIPS2017_7062, lime}.

Most of the recent advancements have prioritized interpretability in image classification tasks, with little development in other important tasks like image segmentation. Image segmentation is an active area of research in medical domains, for tumor and disease identification~\cite{drozdzal2018learning, oktay2018attention}. However, if these tools are to be used in practice by medical professionals to drive complex and life-changing decisions, it is necessary to be able to explain and understand how segmentations are made.

Motivated by these applications, we introduce \textit{U-Noise}, a novel method to interpret segmentation models through noisy image occlusion \footnote{The code for this work can be found at \url{https://github.com/teddykoker/u-noise}}. U-Noise learns to develop interpretability maps from training images, which makes it an extensible and easily deployable interpretation method; once trained, U-Noise is quick to generate interpretability maps for new images.

U-Noise interprets an already trained segmentation model (we name it the Utility model), whose parameters are frozen,  and finds the context that it uses to semantically segment an image to find a target through a systematic process of adding noise. 
For example, in the case of pancreas segmentation in CT-scans, U-Noise finds the context organs used by the model to identify where the pancreas is. 

\begin{figure*}
    \centering
     \includegraphics[width=0.95\linewidth]{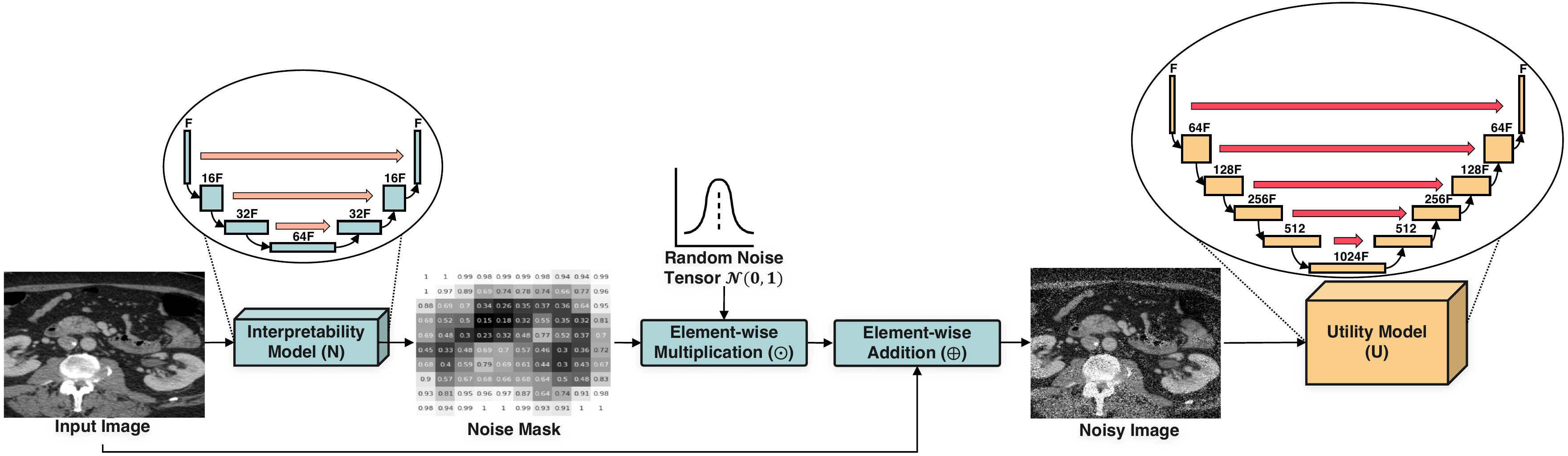}
     \caption{The components and the training process of our proposed U-Noise method. The model we are trying to interpret (utility model, $U$) is frozen, and the interpretability model is being trained. The interpretability model's output is the standard deviation of the noise tolerated by each pixel of the input. It is then multiplied by random noise and fed to $U$, to measure its utility for the loss function, and update the interpretability model accordingly.}
     \label{fig:u-noisediagram}
     \vspace{-1ex}
    \label{fig:prop}
    \vspace{-1ex}
\end{figure*}

U-Noise builds on the intuition that if the model is resilient to the addition of noise with high scale (standard deviation) to a pixel, then that pixel is not important with respect to the model we are trying to interpret. 
We train a small neural network called an interpretability model, to directly parameterize the noise distribution (noise mask) for each pixel in an input image, such that the segmentation accuracy is not harmed.
Once the training is finished, we can generate an "importance map" for each input image by feeding it to the interpretability model. Pixels that would have received noise with a higher scale are less important and vice-versa.
Figure~\ref{fig:u-noisediagram} shows the workflow of U-noise for training the interpretability model. 

There are only a few previous works that focus on segmentation interpretability~\cite{seg-gradcam,hoyer2019grid}. These works solve the problem in a lower resolution setup, and then up-sample to get pixel level explanation. They either return explanations for a part of the image~\cite{seg-gradcam, hoyer2019grid}, a single pixel~\cite{seg-gradcam}, or for the whole class~\cite{seg-gradcam}. Our method on the other hand, seamlessly returns the relative importance of each pixel, with respect to the entire input.

\section{Related Work}

Since our work is on the interpretability of image segmentation, and our proposed method trains noise distributions directly, we divide the related work into two sections: DNN interpretability methods for vision tasks and probability distribution training methods for neural networks. 

\textbf{Interpretability.} CAM~\cite{zhou2016learning}, Grad-CAM~\cite{selvaraju2017grad}, occlusion sensitivity~\cite{zeiler2014visualizing} and T-CAV~\cite{melis2018towards} are the state-of-the-art methods for classification interpretation using CNNs. Grad-CAM uses the gradients of target concepts to measure sensitivity and offer explanations. However, this is done for intermediate activations and upsampled to match the original image, which causes some loss in interpretation power. Occlusion sensitivity covers parts of the input to find important regions and is therefore computationally intensive.  Hoyer et al.~\cite{hoyer2019grid} use perturbations for interpreting segmentation models, however, they do not train the noise parameters as we do, and they also use upsampling for getting pixel-level explanations, unlike our framework which offers importance levels for each pixel directly. 

\textbf{Distribution Training.} 
Mireshghallah et al. introduced Cloak~\cite{mireshghallah2020principled}, a method that trains distributions to obscure parts of the image for privacy purposes. Our work is also related to a method called Bayes by backprop~\cite{blundell2015weight}, which trains distribution parameters using backpropagation, for weights of Bayesian neural networks. 
%
U-Noise builds upon these methods to directly train dynamic noise maps, where the noise parameters can change based on the input image, unlike previous work which offers static distributions. 

\textbf{Segmentation.}
U-Net~\cite{ronneberger2015u} is a powerful image segmentation architecture which is used extensively in medical domains~\cite{oktay2018attention}. The model consists of a contracting path of convolutions to capture context from the image, followed by an expanding path of deconvolutions to allow precise object localization. The architecture is fast and produces high-resolution segmentations.

\section{U-Noise}
%
%
%
%

%
%
%

%
We build on the intuition that if a pixel is important for the decision making of a trained model that is supposed to run a task, in our case segmentation, then adding noise with high scale to this pixel would harm that model's utility.
As such, we define a utility model $U$ (with parameters $\theta_U$), which is the model whose decisions we want to interpret, and an interpretability model $N$ (with parameters $\theta_N$, whose purpose is to compute the scale of noise to be applied to each pixel, while simultaneously maximizing utility model performance, as well as the total scale of noise applied.
In the segmentation task explored in this work, both the $U$ and $N$ have a U-Net architecture.
Figure~\ref{fig:u-noisediagram} shows the components of U-Noise and how they interact.

Given an image $x \in \mathbb{R}^{C \times H \times W}$ with $C$ channels, and a height and width of $H$ and $W$ pixels respectively, the noise model produces a mask $B \in \mathbb{R}^{H \times W} = N(x; \theta_N)$ contained in the range $[0, 1]$ by the sigmoid function. $B$ is then linearly scaled to range $[\sigma_\text{min}, \sigma_\text{max}]$ (hyperparameters) to determine the scale of noise applied to each pixel. The noised image, $x'$, is then produced:
\begin{equation}
    x'_{ij} = x_{ij} + B_{ij}(\sigma_\text{max} - \sigma_\text{min})\epsilon + \sigma_\text{min}
\end{equation}

%
Where $\epsilon \sim \mathcal{N}(0, 1)$ is sampled from the standard normal distribution. The predicted segmentation mask, $\hat{y} = U(x'; \theta_U)$ is then computed using the \textit{frozen} utility model. The goal of the training process for the interpretability model is:
\begin{equation*}
    \min_{\theta_N} \quad- \log \Pr(\hat{y} |y;\theta_U, \theta_N) - \lambda\log B
\end{equation*}

Where $\theta_N$ and $\theta_U$ are the parameters of the interpretability and utility model, respectively. The first term is the utility loss, which is the cross-entropy loss that incentivizes $N$ to orchestrate the noise such that the utility of the main model is unharmed. 
The second term is $-\log{B}$, which incentivizes the growth of the scale (standard deviation) of the noise standard deviation for each pixel so that we can find the unimportant ones. Algorithm~\ref{alg:u-noise} shows the training process in U-Noise.
The trade-off between the two terms is governed by $\lambda$, the \textit{noise ratio} hyperparameter.

Training of the interpretability model only takes place once, after which one only needs to run inference on the interpretability model to interpret a new image.

\begin{algorithm}
 \caption{Train U-Noise}
 \label{alg:u-noise}
 
  \KwIn{Input images $X$, segmentation masks $Y$, learning rate $\eta$, noise model $N$ with parameters $\theta_N$, trained utility model $U$ with parameters $\theta_U$, learning rate $\eta$, min/max noise scale $\sigma_\text{min}$ and $\sigma_\text{max}$
  }
  
  \KwOut{Trained noise model parameters $\theta_N$}
  \While{not converged}{
    \For{$x \in X, y \in Y$} {
        $B \leftarrow N(\theta_N, x)$ \;
        $\epsilon \sim \mathcal{N}(0, 1)$ \;
        $x' \leftarrow x + B(\sigma_\text{max} - \sigma_\text{min})\epsilon + \sigma_\text{min}$ \;
        $\hat{y} \leftarrow U(x';\theta_U)$ \;
        $\mathcal{L}_\text{utility} \leftarrow \text{CrossEntropy}(y, \hat{y})$ \;
        $\mathcal{L} \leftarrow \mathcal{L}_\text{utility} - \lambda\log B$\;
        $\theta_N \leftarrow \theta_N + \eta \frac{d\mathcal{L}}{d\theta_N}$\;
    } 
  } 
\end{algorithm}

\section{Results}

\subsection{Experimental Setup}

We apply the U-Noise framework to a pancreas segmentation task on a dataset  from~\cite{decathlon}. This publicly available dataset contains CT scans of the pancreatic region and associated masks outlining the pancreas.
%
For all experiments, we use a pre-trained U-Net model as the \textit{Utility} model.
%
We train various sizes of U-Noise interpretability models, outlined in Table~\ref{table:noise}. Each model is trained with a minibatch size of 8 for 100 epochs.

%


\begin{table}[bth]
\centering
  \resizebox{0.85\columnwidth}{!}{
    \begin{tabular}{|c | c | c | c |}
      \hline
      Model name & Depth & Channels & Parameter count \\
      \hline
      Utility & 5 & $2^6$ & 34M \\
      U-Noise Small & 2 & $2^4$ & 10K \\
      U-Noise Medium & 3 & $2^4$ & 130K \\
      U-Noise Large & 4 & $2^4$ & 537K \\
      \hline
    \end{tabular}
  }
  \caption{Details of the Utility and U-Noise models used in experiments. \textit{Depth} denotes the number of downsample and upsample layers; \textit{Channels} denotes the number of output channels of the first layer.}
  \label{table:noise}
\end{table}

\subsection{Method Comparison}

\begin{figure}[tb]
\resizebox{0.9\columnwidth}{!}{
\begin{tabular}{c@{}c@{}c@{}c}
Image + Mask & U-Noise Large & Occlusion & Grad-CAM \\
\includegraphics[width = .125\textwidth]{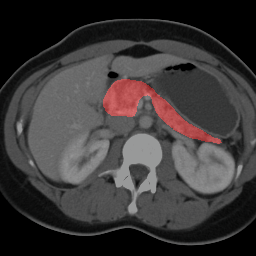} &
\includegraphics[width = .125\textwidth]{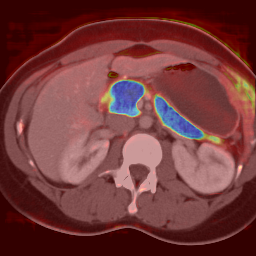} &
\includegraphics[width = .125\textwidth]{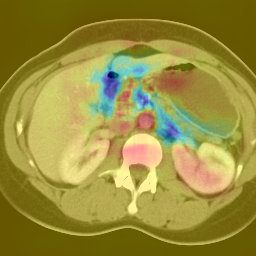}  &
\includegraphics[width = .125\textwidth]{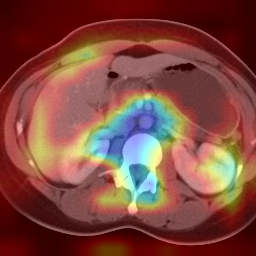}\\
\includegraphics[width = .125\textwidth]{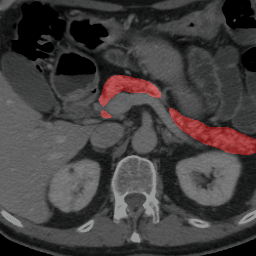} &
\includegraphics[width = .125\textwidth]{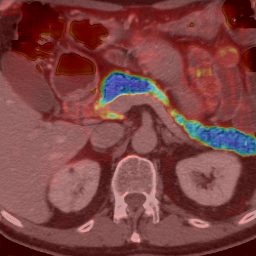} &
\includegraphics[width = .125\textwidth]{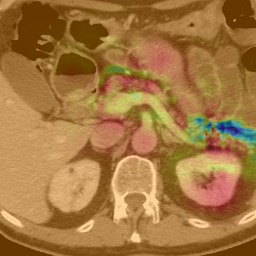}  &
\includegraphics[width = .125\textwidth]{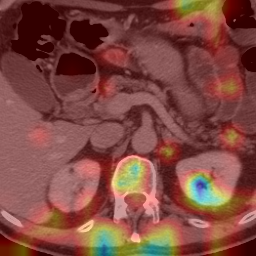}\\
\includegraphics[width = .125\textwidth]{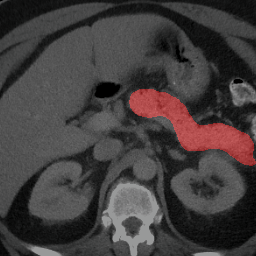} &
\includegraphics[width = .125\textwidth]{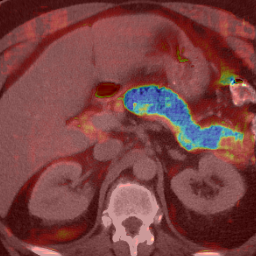} &
\includegraphics[width = .125\textwidth]{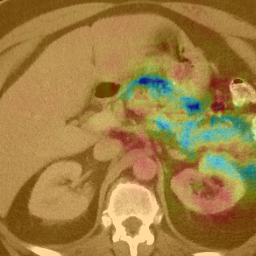}  &
\includegraphics[width = .125\textwidth]{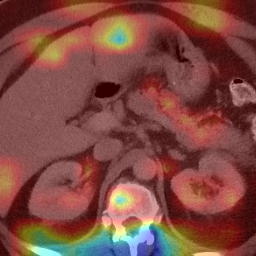}\\
\includegraphics[width = .125\textwidth]{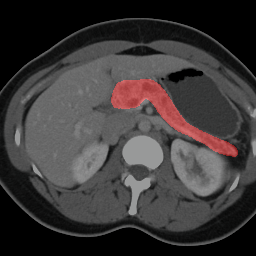} &
\includegraphics[width = .125\textwidth]{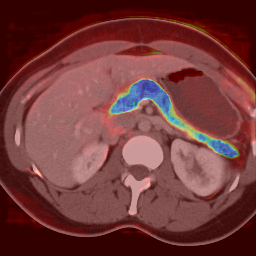} &
\includegraphics[width = .125\textwidth]{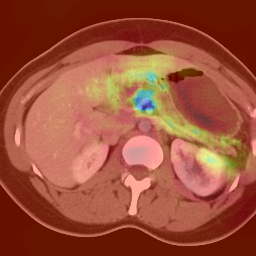}  &
\includegraphics[width = .125\textwidth]{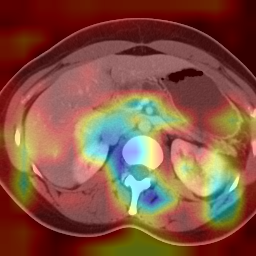}\\
\end{tabular}
}
\caption{Comparison of U-Noise Large model to Occlusion Sensitivity and Grad-CAM interpretation techniques.}
\label{fig:method-comparison}
\end{figure}

Fig.~\ref{fig:method-comparison} plots a random selection of images from the Pancreas-CT validation set, interpreted by different methods: U-Noise Large, occlusion sensitivity, and Grad-CAM. For U-Noise, we visualize the positive values retrieved prior to the sigmoid layer in the noise model. For occlusion sensitivity, we convolve a 15x15 pixel window through the image with a stride of 2 pixels. At each position, we report the dice coefficient relative to the unoccluded image. Finally, for Grad-CAM, we follow \cite{seg-gradcam, melis2018towards} and obtain a heatmap with respect to the convolutional layer at the bottleneck of the Utility model. 

Qualitatively, it can be seen that, unlike the other methods, U-Noise places the highest importance on the pancreas itself. Logically, this can be interpreted as a pixel being labeled as part of the pancreas increases the probability of neighboring pixels also being labeled as the pancreas. Other pixels of importance tend to be on the borders of organs near the pancreas and the outside edges of the body, demonstrating what the model might be using to signpost towards the pancreas. In comparison, it is not clear how to interpret the heatmaps from the other techniques.

Fig.~\ref{fig:threshold-comparison} shows pixels at each decile of importance, as ranked by the interpretability model, for an example image by setting a threshold for occlusion map value below which pixels were zeroed. The first pixels to be un-zeroed (the most important) relate to the pancreas. Next are the edges of the body and nearby organs, suggesting that these are used to identify pancreas location and extent.

\subsection{Utility Tradeoff}

\begin{figure}[tb]
\centering
\begin{subfigure}{.45\textwidth}
\includegraphics[width=\hsize]{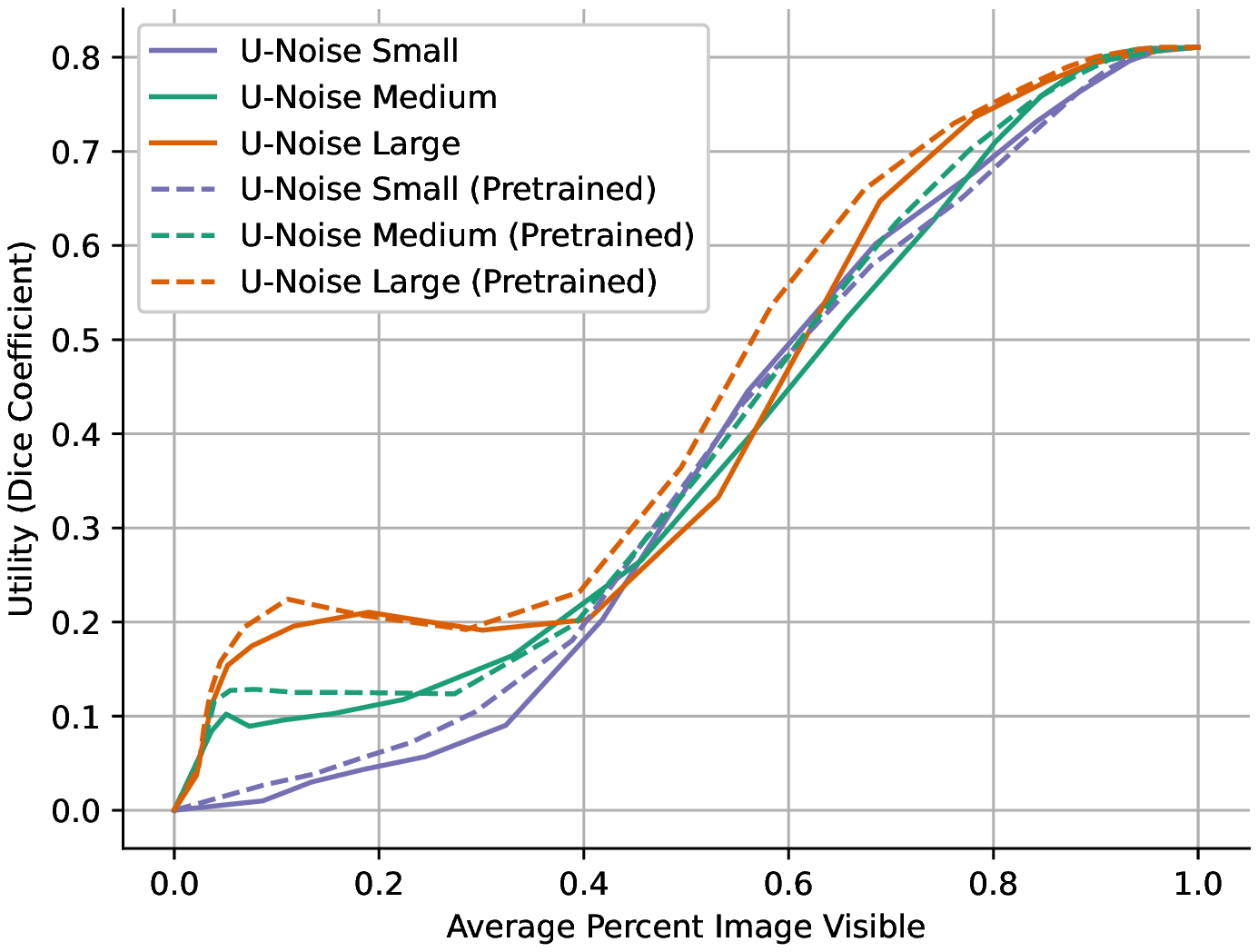}
\caption{Percent image occlusion vs utility performance}
\label{fig:zerotradeoff}
\end{subfigure}
\begin{subfigure}{.45\textwidth}
\resizebox{0.95\columnwidth}{!}{
\begin{tabular}{c@{}c@{}c@{}c}
\includegraphics[width = .25\hsize]{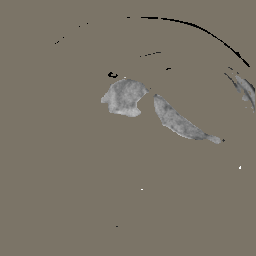} &
\includegraphics[width = .25\hsize]{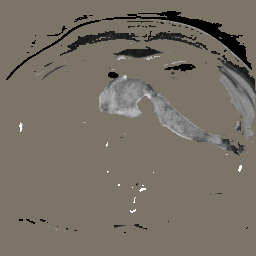} &
\includegraphics[width = .25\hsize]{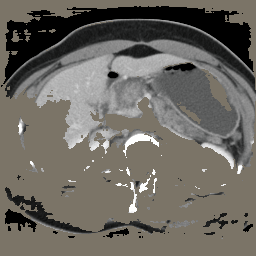}  &
\includegraphics[width = .25\hsize]{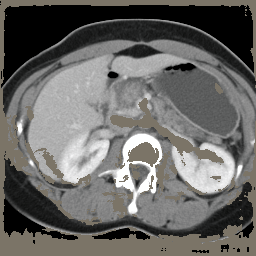}\\
\end{tabular}
}
\caption{Image thresholded at $B \leq [0.1, 0.3, 0.5, 0.7]$}
\label{fig:threshold-comparison}
\end{subfigure}

\caption{Effect of thresholding images to the percentage of the image visible, and the accuracy of the downstream utility model. Models marked \textit{Pretrained} were first trained on the pancreas segmentation task before the noise task. Intuitively, we observe a direct relationship between the percent of the image visible and the ability of the utility model to segment the image. }
\end{figure}

Fig.~\ref{fig:zerotradeoff} plots the dice coefficient of U-Noise against the percentage of zeroed pixels in an image. The dice coefficient decreases minimally until less than $~90\%$ of the image is visible, after which the model utility decreases steadily. This suggests that the utility model requires context from a large number of pixels in the image to make the segmentation.

The differences in utility response of the models when up to 50\% of the image is visible suggests that the pixel priority learned by the large model is sub-optimal.

We analyze the importance assigned to each pixel by the occlusion map by zeroing pixels with an occlusion map value above a certain threshold. Fig.~\ref{fig:zerotradeoff} plots the impact on dice coefficient against the zero threshold and the corresponding percentage pixels in the image which have been zeroed.

\subsection{Model Pretraining}

\begin{figure}[tb]
\centering
\includegraphics[width=.45\textwidth]{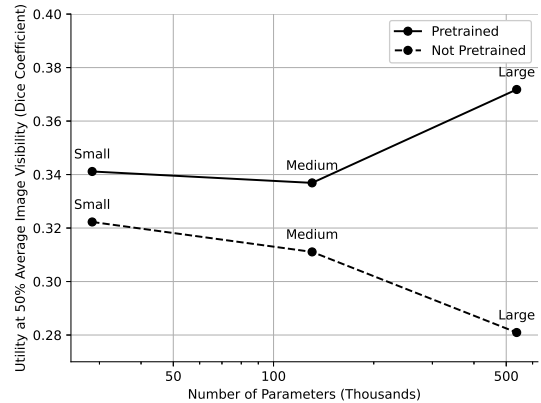}
\caption{Effect of model size and pretraining. We observe that the Small model shows the least improvement to pretraining on the original task, while it seems necessary for the Large model.}
\label{fig:pretraining-model-size}
\end{figure}

Fig.~\ref{fig:pretraining-model-size} plots the dice coefficient of U-Noise when 50\% of the image is visible, both when the interpretability model was and was not initialized with the trained Utility model parameters. Unsurprisingly, U-Noise performs better with a pre-trained interpretability model as it takes advantage of the pancreas identification learned by the Utility model. The inverse relationship between interpretability model size and U-Noise utility suggests that the interpretability model struggles to simultaneously learn the implicit tasks of pancreas identification and pixel importance ranking.

Although the absolute differences in dice coefficient are small, taken to its extreme this may suggest that large interpretability models must be pre-trained in order for U-Noise interpretations to work well. In non-segmentation tasks, where the interpretability and Utility models may not share an architecture and therefore cannot easily share parameters, it may not be possible to use large interpretability models. 

\subsection{Run-time Analysis}

\begin{table}[bt]
  \resizebox{\columnwidth}{!}{
    \begin{tabular}{|c | c | c | c |}
      \hline
      Method & Inference Time (seconds) & Training Time & Parameters \\
      \hline
      U-Noise Large & 0.0031 & 4 hours 56 min. 30 sec. & 537K \\
      Grad-CAM & 0.046 & NA & 34M \\
      Occlusion Sensitivity & 201.37 & NA & 34M \\
      \hline
    \end{tabular}
  }
  \caption{Compute requirements of different interpretability methods. Average over 10 trials. Methods compared using an NVIDIA 2070 Super GPU.}
  \label{table:overhead}
\end{table}

Table~\ref{table:overhead} compares the computational cost of U-Noise to Grad-CAM and Occlusion Sensitivity.
Once trained, our method is orders of magnitude quicker to infer than the others, making it more applicable for use at scale. In situations where specialized, high-performance compute is not easily available (for example in hospitals, for use-cases explored in this work), U-Noise remains a useful interpretation method.

\section{Conclusion and Future Work}

In this paper, we present U-Noise, a noise-based framework for interpreting image segmentation models. U-Noise discovers the pixels on which the segmentation model relies to makes its decision, by applying additive noise and observing how sensitive the model utility is to changes in the value of each pixel. While this work has only considered U-Noise in segmentation tasks, the method can in principle be applied to other tasks.

Additionally, the architecture presented in this work offers a lightweight method for occluding sections of an image without severely compromising task utility. Therefore, U-Noise could be utilized to obscure user data on-device in a more directed way than existing privacy methods, such as local differential privacy \cite{dwork2006differential}. The efficacy of U-Noise for the purpose of user privacy should be explored in detail.

\textbf{Ethical Considerations.}
We have made sure that the data we used was collected under HIPAA provenance and that the privacy of the patients who have contributed to the dataset is not violated.

\bibliographystyle{IEEEbib}
\bibliography{refs}

\end{document}